\documentclass[pdflatex,sn-basic]{sn-jnl}

\usepackage{graphicx}%
\usepackage{multirow}%
\usepackage{xspace}
\usepackage{amsmath,amssymb,amsfonts}%
\usepackage{amsthm}%
\usepackage{mathrsfs}%
\usepackage[title]{appendix}%
\usepackage{xcolor}%
\usepackage{textcomp}%
\usepackage{url}
 \usepackage{hyperref}
 \usepackage{wrapfig}
 \usepackage{amssymb}
 \usepackage{amsmath}
\usepackage{manyfoot}%
\usepackage{booktabs}%
\usepackage{bbding}
\usepackage{pifont}
\usepackage{algorithm}%
\usepackage{algorithmicx}%
\usepackage{algpseudocode}%

\makeatletter
\DeclareRobustCommand\onedot{\futurelet\@let@token\@onedot}
\def\@onedot{\ifx\@let@token.\else.\null\fi\xspace}
\def\eg {\emph{e.g}\onedot} 
\def\ie{\emph{i.e}\onedot} 
 
\def\etc{\emph{etc}\onedot}

\makeatother

\theoremstyle{thmstyleone}%
\theoremstyle{thmstyletwo}%

\theoremstyle{thmstylethree}%

\begin{document}

\title[Auditing Data Provenance in Real-world Text-to-Image Diffusion Models for Privacy and Copyright Protection]{Auditing Data Provenance in Real-world Text-to-Image Diffusion Models for Privacy and Copyright Protection}


\author[1,2]{\fnm{Jie} \sur{Zhu}}\email{zhujie@stu.pku.edu.cn}

\author*[1,2]{\fnm{Leye} \sur{Wang}}\email{leyewang@pku.edu.cn}


\affil[1]{\orgdiv{Key Lab of High Confidence Software Technologies (Peking University)}, \orgname{Ministry of Education}, \orgaddress{\postcode{100871}, \state{Beijing}, \country{China}}}

\affil[2]{\orgdiv{School of Computer Science}, \orgname{Peking University}, \orgaddress{\postcode{100871}, \state{Beijing}, \country{China}}}



\abstract{Text-to-image diffusion model since its propose has significantly influenced the content creation due to its impressive generation capability. However, this capability depends on large-scale text-image datasets gathered from web platforms like social media, posing substantial challenges in copyright compliance and personal privacy leakage. Though there are some efforts devoted to explore approaches for auditing data provenance in text-to-image diffusion models, existing work has unrealistic assumptions that can obtain model internal knowledge, \eg, intermediate results, or the evaluation is not reliable. To fill this gap, we propose a completely black-box auditing framework called \textbf{F}eature \textbf{S}emantic \textbf{C}onsistency-based \textbf{A}uditing (FSCA). It utilizes two types of semantic connections within the text-to-image diffusion model for auditing, eliminating the need for access to internal knowledge. To demonstrate the effectiveness of our FSCA framework, we perform extensive experiments on LAION-mi dataset and COCO dataset, and compare with eight state-of-the-art baseline approaches. The results show that FSCA surpasses previous baseline approaches across various metrics and different data distributions, showcasing the superiority of our FSCA. Moreover, we introduce a recall balance strategy and a threshold adjustment strategy, which collectively allows FSCA to reach up a user-level accuracy of 90\% in a real-world auditing scenario with only 10 samples/user, highlighting its strong auditing potential in real-world applications. Our code is made available at \url{https://github.com/JiePKU/FSCA}.}

\maketitle

\section{Introduction}


Recently, text-to-image diffusion models~\cite{rombach2022high, saharia2022photorealistic, zhumole}, such as Stable Diffusion~\footnote{\url{https://stability.ai/}} and Midjourney~\footnote{\url{https://www.midjourney.com/home}} has redefined how to leverage information technology (IT) for content creation~\cite{daugherty2008exploring, dale2014content, vohra2019customer}. These models, trained on extensive datasets of text-image pairs, enable the automated generation of high-quality visuals from textual descriptions, streamlining creative workflows and transforming content creation from traditional manual processes to highly automated, AI-driven operations. This shift exemplifies the growing integration of advanced AI capabilities into enterprises, reshaping creative industries such as advertising and digital design. 

However, the impressive generative capabilities of text-to-image models rely heavily on large-scale, high-quality text-image datasets, often sourced from web platforms such as social media. This data collection practice introduces significant challenges in copyright compliance and individual privacy~\footnote{\url{https://adguard.com/en/blog/ai-personal-data-privacy.html?utm_source=chatgpt.com}}. 
For instance, Getty Images has filed lawsuits against Stability AI, the developer of Stable Diffusion, alleging unauthorized copying and processing of millions of copyrighted images~\footnote{\url{https://www.reuters.com/legal/getty-images-lawsuit-says-stability-ai-misused-photos-train-ai-2023-02-06/}}. 
These issues highlight the critical importance of implementing robust mechanisms to audit the images used in text-to-image model training, ensuring adherence to copyright regulations~\cite{flew2013copyrights, garnham2005cultural}. Additionally, as users become increasingly aware of privacy concerns~\cite{lwin2007consumer, rong2022social, cao2024consequences}, they seek assurances that their personal data has not been exploited without consent~\cite{ke2023privacy, aubert2022privacy}. This growing awareness also emphasizes the urgency of developing effective auditing approaches for public use.

Prior to this work, researchers have explored approaches for auditing text-to-image diffusion models, aiming to assist organizations and individual users in identifying whether their images (and corresponding text information) are used for model training without authorization. These approaches 
can be broadly categorized into metric-based approaches~\cite{matsumoto2023membership, li2024unveiling, duan2023diffusion, kongefficient, fu2023probabilistic, dubinski2024towards} and classifier-based approaches~\cite{zhai2024membership, wu2022membership}. Metric-based approaches leverage pre-defined metrics, such as loss values~\cite{duan2023diffusion, kongefficient} and image structural similarity~\cite{li2024unveiling, wang2004image}, to evaluate the membership status~\footnote{For simplicity, we adopt the terms "member" and "non-member" from previous studies~\cite{shokri2017membership, hu2022membership, zhu2024unified, zhu2022safety, zhu2024safety} in the field of membership inference to refer to images used in training and those that are not, respectively. We will give a detailed explanation why we adopt these terms from the field of membership inference in Sec~\ref{sec:membership inference}.} of a given text-image pair. 
Classifier-based approaches extract features from a given text-image pair and feed the features into a trained classifier such as a neural network. Subsequently, this classifier yields probability scores to assess the likelihood that this text-image pair is from the training dataset.

\begin{figure}[t]
\centering\centerline{\includegraphics[width=1.0\linewidth]{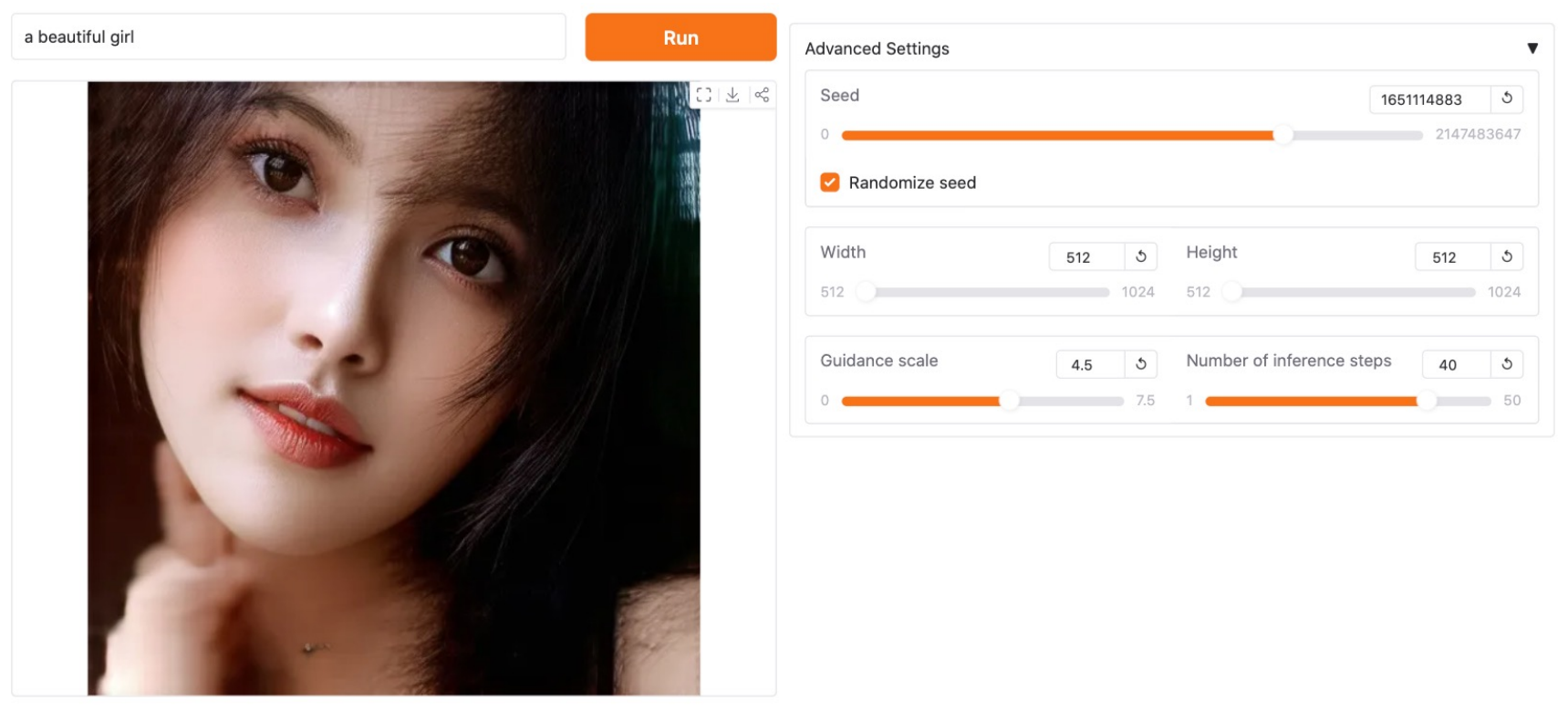}}
\caption{A screenshot of a text-to-image diffusion model demo on a website.}
\label{fig:demo-case}
\end{figure}

While these studies have proposed auditing approaches to address the issue, most have been validated under \textit{unrealistic} gray-box setting~\cite{matsumoto2023membership, duan2023diffusion, kongefficient, fu2023probabilistic, li2024unveiling, zhai2024membership}. In this setting~\footnote{Note that in the real-world scenarios, the requirements of auditing in gray-box and white-box settings are nearly
identical. We use gray-box setting here for consistency with previous works~\cite{duan2023diffusion, kongefficient, zhai2024membership}}, users are accessible to the visual and text encoders of the target text-to-image diffusion model and can extract intermediate outputs, such as embeddings of given images, text prompts, and denoised latent. Unfortunately, this setting is rarely valid in real-world scenarios. As illustrated in Fig~\ref{fig:demo-case}, which is a website demo of a deployed real-world text-to-image diffusion model in a system (text-to-image diffusion system~\footnote{
In the rest of the paper, without incurring the ambiguity, we will use ‘text-to-image diffusion system’ to indicate the ‘a deployed real-world text-to-image diffusion model in a system’ for clarity.}), users typically have only black-box access. In this setting, users can generate images from input text and adjust parameters like inference steps but cannot access the text-to-image diffusion system’s intermediate outputs or model weights. This limitation renders most of the existing auditing approaches practically unsuitable. Recognizing this gap, researchers start to consider black-box auditing approaches~\cite{dubinski2024towards, wu2022membership}. However, the performance is either subpar due to an overly simplistic inference metric (reconstruction pixel error)~\cite{dubinski2024towards} or unreliable as members and non-members in the evaluation benchmarks are from different datasets (distributions), causing the auditing model to function more like a dataset classifier,  \ie, identifying which distribution a data sample belongs to, rather than a dependable auditing tool~\cite{wu2022membership}. Therefore, it is still crucial to develop effective and reliable black-box auditing approaches. 


\begin{figure}[t]
\centering\centerline{\includegraphics[width=1.0\linewidth]{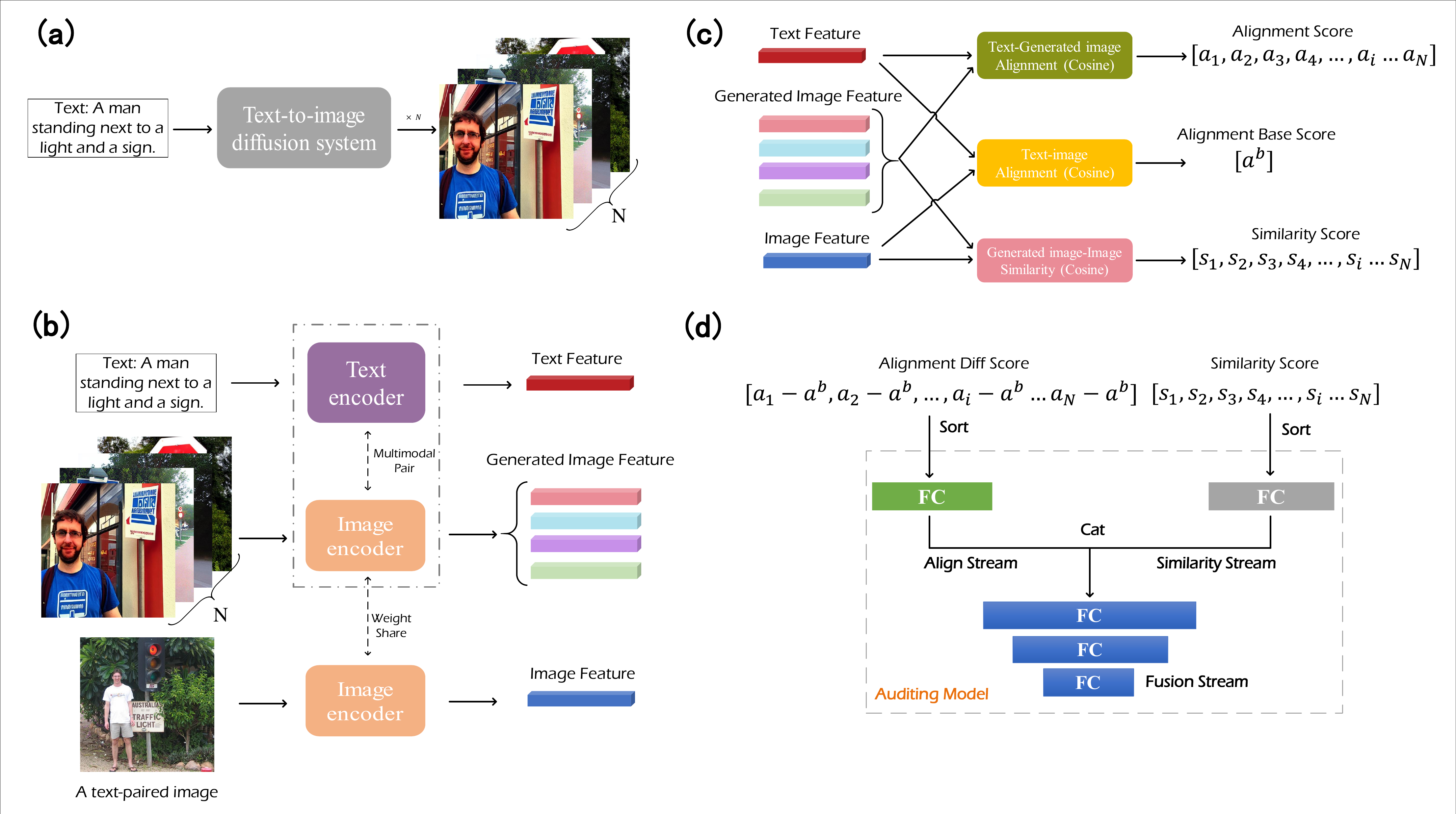}}
\caption{An overview of our feature semantic consistency-based auditing framework (\textbf{FSCA}). As illustrated, our FSCA is a completely black-box auditing framework as we do not access the internal knowledge of a target text-to-image diffusion system and utilizes the generated images only. }
\label{fig:overview}
\end{figure}

Against this backdrop, this work aims to develop such approach, allowing users to verify their data involvement in model training while advancing the transparency and accountability of text-to-image diffusion systems. Computational design science provides researchers with the guideline that the IT artifact’s design should be inspired by key domain \textit{characteristics}~\cite{nunamaker1990systems, hevner2004design, gregor2013positioning}, from which our approach benefits. Our key insight is that training a text-to-image diffusion model essentially involves bridging the \textit{connection} in text-image pairs and modeling their distribution.
\begin{wrapfigure}{r}{0.4\textwidth}
    \centering
    \vskip -0.2in
\includegraphics[width=0.9\linewidth]{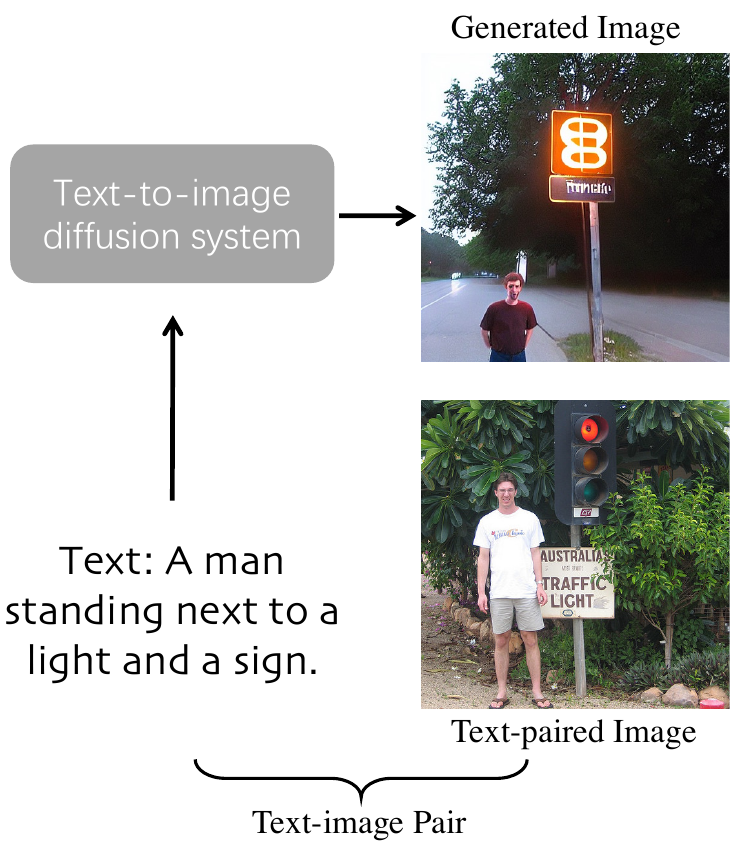}
 \vskip -0.05in
    \caption{The concept illustration of text-image pair, text-paired image, and generated image.}
    \vskip -0.15in
    \label{fig:concept}
\end{wrapfigure}
Therefore, we can inversely leverage this connection (characteristic) for auditing without requiring access to internal knowledge. Specifically, we consider following two manners to capture this connection. 
If a text-image pair is used during training:

$\bullet$ First, the generated image of that text as illustrated in Fig~\ref{fig:concept} is likely to be semantically similar to the text-paired image; 

$\bullet$ Second, the generated image typically aligns with the given text to a degree similar to that of the original text-image pair~\footnote{In fact, we could incorporate another manner—text-text semantic similarity—by utilizing a caption model to describe the content of generated images and then measuring the similarity between the original and generated text. However, since this manner relies on a caption model and is therefore more indirect than the two manners above, we have opted not to adopt it in this work, leaving it for future exploration.}. 

In particular, scarcity theory~\cite{mullainathan2013scarcity} and information theory~\cite{shannon1948mathematical} suggest that  resources (or information) derive greater value from their rarity. Similarly, in the context of text-to-image diffusion generation, when the text includes rare words or describes uncommon scenes, the text-image connection may become more distinctive, making the aforementioned two manners more pronounced.

To the end, we develop a black-box and practical feature-level semantic-guided auditing framework called \textbf{F}eature \textbf{S}emantic \textbf{C}onsistency-based \textbf{A}uditing (\textbf{FSCA}) as shown in Fig~\ref{fig:overview}. FSCA explicitly establishes connections between the given text, original image, and generated image by focusing on semantic consistency. Specifically, we first generate $N$ images with text-to-image diffusion system in Fig~\ref{fig:overview} (a). Then in Fig~\ref{fig:overview} (b) we extract abstract features from text and (generated) images using multimodal text and image encoders (\eg, CLIP~\cite{radford2021learning}), respectively.  Fig~\ref{fig:overview} (c) computes feature-level semantic similarity and alignment to evaluate semantic consistency. The resulting scores, referred to as membership features~\cite{zhu2024unified}, are used as input for an auditing model as shown in Fig~\ref{fig:overview}~(d).

To verify the effectiveness of the proposed framework, we rigorously evaluate our FSCA framework against previous baseline approaches on two widely-used and representative datasets including LAION-mi~\cite{dubinski2024towards} and COCO~\cite{lin2014microsoft} following previous work~\cite{zhai2024membership}. LAION-mi is sampled from LAION-5b~\cite{schuhmann2022laion} that contains billions of image-text pairs to facilitate the development of industry-level multi-modal products. Hence, LAION-mi is a more \textit{realistic} dataset. Moreover, through elaborate processing, LAION-mi ensures the distribution consistency between member/non-member sets, enabling more reliable evaluation results. COCO is sampled from MS-COCO~\cite{lin2014microsoft} that serves as a diverse academic benchmark for visual understanding and generation. In our experiments on LAION-mi, FSCA outperforms all black-box and gray-box approaches, showing its great potential of auditing. In our experiments on COCO, FSCA produces superior performance over all black-box approaches and outperforms most gray-box approaches. Additionally, we introduce a recall balance strategy and a threshold adjustment strategy, enabling FSCA to reach up a user-level accuracy of 90\% in a real-world user-level auditing scenario with only 10 samples/user.

Our contribution can be summarized as follows:

$\bullet$ We propose a completely black-box feature-level semantic-guided consistency-based auditing framework called FSCA. In brief, it calculates the semantic consistency of features from text and (generated) images and collect
them as membership feature to perform auditing.

$\bullet$ We evaluate the proposed FSCA framework against previous black-box and gray-box baseline approaches on LAION-mi and COCO, confirming its effectiveness and generalization. Moreover, FSCA can reach up a user-level accuracy of 90\% in a real-world auditing scenario, showing its great potential of auditing in realistic application.

$\bullet$ FSCA effectively matches and even surpasses gray-box-based membership auditing approaches, marking a significant step toward real-world black-box auditing in text-to-image diffusion systems.




\subsection{Related Work}


\subsubsection{Text-to-image Diffusion Generation}
Since the introduction of text-to-image diffusion generation~\cite{rombach2022high, saharia2022photorealistic, zhumole}, researchers have made significant advancements across various interconnected domains, including model architecture, multimodal alignment, and efficiency. Model architecture and multimodal alignment directly influence how effectively text prompts are translated into coherent and visually compelling images. Recent advancements in architecture and diffusion space, such as from hybrid transformer-convolutional designs~\cite{rombach2022high} to pure transformer-based designs~\cite{chenpixart} and from the original pixel space~\cite{saharia2022photorealistic} to a compressed latent space~\cite{rombach2022high}, enables more efficient representation learning. 
Multimodal alignment ensures the alignment between generated images and textual descriptions~\cite{podellsdxl, esser2024scaling} by leveraging large-scale, text-image datasets such as LAION~\cite{schuhmann2022laion} and integrating pretrained text encoders like CLIP~\cite{radford2021learning} and T5~\cite{raffel2020exploring} to extract text semantic. Efficiency in training and inference is another critical area of research, particularly given the resource-intensive nature of text-to-image diffusion models~\cite{ho2020denoising}. Methods such as improved noise scheduling~\cite{nichol2021improved, karras2022elucidating}, progressive training~\cite{salimansprogressive, song2023consistency}, and superior sampler~\cite{songdenoising, lu2022dpm} have substantially reduced computational overhead.
Differing from from these studies, our work focuses on the privacy and copyright risks involved in text-to-image diffusion models, particularly due to their large-scale data collection behaviors on social media. 





\subsubsection{Membership Auditing}

Membership auditing is developed to auditing whether a data is used for training a neural model~\cite{shokri2017membership, salem2018ml}. It is initially proposed for classification models~\cite{hu2022membership, shokri2017membership, salem2018ml} and afterward generalize to models of various tasks~\cite{nasr2019comprehensive, he2020segmentations, wen2024membership, duddu2020quantifying, zhai2024membership} such as graph~\cite{duddu2020quantifying}, federated learning~\cite{nasr2019comprehensive}, and generation~\cite{zhai2024membership} due to the importance of protecting individual privacy and intellectual property. In particular, likely driven by the impressive generation quality of text-to-image diffusion models, with their advancement and widespread adoption, they have garnered increasing concerns from researchers.
Currently, researchers have primarily developed two types of membership auditing approaches including metrics-based approaches~\cite{matsumoto2023membership, li2024unveiling, duan2023diffusion, kongefficient, fu2023probabilistic, dubinski2024towards} (e.g., loss value) and classifier-based approaches~\cite{zhai2024membership, wu2022membership}. However, most approaches operate in gray-box settings, assuming an unrealistic level of access to real-world text-to-image diffusion systems. Though researchers start to consider more realistic black-box auditing approaches~\cite{dubinski2024towards, wu2022membership}. the performance is either poor~\cite{dubinski2024towards} or unreliable~\cite{wu2022membership} due to unreasonable design and evaluation. As a result, it is still necessary to develop a more practical and effective black-box membership auditing approach tailed for a text-to-image diffusion system.

\section{Results}

In this section, we first introduce two potential experiment settings about user knowledge in auditing. Then we demonstrate the superiority of our FSCA framework compared to previous approaches under two completely different user knowledge settings on LAION-mi and COCO datasets. LAION-mi holds greater real-world relevance as it is derived from LAION-5b, a dataset widely adopted for developing industry-grade applications and products, while the COCO dataset is used as a diverse academic benchmark for visual understanding and generation. Additionally, LAION-mi ensures distribution consistency between member and non-member sets for more reliable evaluation results. Hence we pay more attention to the results on LAION-mi. We also evaluate the performance of our FSCA framework under various variations, \eg, the type of membership feature and query number $N$, and compare the performance of one-branch and two-branch structure of an auditing model. Finally, we consider a real-world user-level auditing experiment to further show the application value of our FSCA in practical auditing.

\subsection{User Knowledge in Auditing}
Let' s consider a user who wants to audit whether her images in social media are used for training a text-to-image diffusion model such as Stable Diffusion but \textit{only} has black-box access to this text-to-image diffusion system. In this context, following previous studies~\cite{liu2022ml, zhu2024unified}, we primarily consider two kinds of scenarios for the user knowledge including partial setting and shadow setting.

\textbf{Partial Setting.} In some fortunate situations, the user might possess partial knowledge of the training dataset of this text-to-image diffusion system, e.g., 50\%. For instance, certain products come with technical reports that might mention the datasets they utilize. We denote this situation as \textit{partial setting}. In this setting, we assume that the user can directly use the partial data to train an auditing model and then auditing his own data. For the sake of the performance comparison, we use the remaining data for evaluation. We use this setting for evaluation if not specified.

\textbf{Shadow Setting.} Besides partial setting, the user often has no knowledge of the training data used for this text-to-image diffusion system. This scenario further restricts the user's prior knowledge and increases the challenges of conducting an auditing. We denote this situation as \textit{shadow setting}.  In this setting, the user has to resort to public datasets, e.g., COCO. Specifically, the user can first fine-tune an open-sourced text-to-image diffusion model with a public dataset and then use this model and the public dataset to train an auditing model. Afterward, the user uses this auditing model to audit her own data. For the sake of the performance comparison, we use COCO as the public dataset and use LAION-mi for evaluation.

\subsection{A Comprehensive Comparison of FSCA}

To demonstrate the effectiveness of our FSCA framework, we conduct a comprehensive comparison with previous approaches in three tasks including LAION-mi in partial setting, LAION-mi in shadow setting, and COCO in partial setting. For all experiments, following previous work~\cite{zhu2022safety, zhu2024unified, duan2023diffusion, kongefficient, fu2023probabilistic, zhai2024membership} we pay more attention on the accuracy (Acc) and report precision (Pre), recall (Rec), F1-score (F1), AUC, and the True Positive Rate (TPR) when the
False Positive Rate (FPR) is 1\% (\i.e., TPR@1\%FPR) for comprehensive comparisons. 

\setlength{\tabcolsep}{0.1cm}{\begin{table*}[htbp]
\centering
\vskip -0.1in
        \caption{Comparison of FSCA with previous approaches on \textbf{LAION-mi in partial setting}. }
			\begin{tabular}{l c c c c c c}
				\toprule
				Approach & Acc$\uparrow$ & Pre$\uparrow$ & Rec$\uparrow$ &  F1$\uparrow$ & AUC$\uparrow$ & TPR@1\%FPR$\uparrow$ \\ 
				\midrule
                    \midrule
                    \multicolumn{4}{l}{\emph{Gray-box baseline}:}\\
				Loss~\cite{matsumoto2023membership} &  53.07 & 55.60 & 30.50 & 39.39 &  53.36  & 1.15  \\
                    PIA~\cite{kongefficient} & 52.35 &  54.85 & 26.55 &  35.78 & 52.59 & 1.00 \\
                    SecMI~\cite{duan2023diffusion} & 52.67  & 52.34 &  59.85 & 55.84 & 52.98 & 0.70 \\
                    PFAMI~\cite{fu2023probabilistic} & 51.60 & 52.91 & 29.05 & 37.51 & 51.46 & 1.25  \\
                    M. C.~\cite{zhai2024membership} &  52.75 & 56.96 & 22.50 & 32.25 & 53.33   & 1.35  \\
    			  CLiD~\cite{zhai2024membership} & 57.40 & 58.52 & 50.90 & 54.44 & 60.58 &  3.05 \\
				\midrule 
    \multicolumn{4}{l}{\emph{Black-box baseline}:} \\
    Pixel Error~\cite{dubinski2024towards} & 50.78 & 51.11 & 35.55 & 41.93 & 49.49 & 0.85 \\
    Attack-IV~\cite{wu2022membership} & 59.72 & 59.13 & 62.42 & 60.73 & 65.64 & 3.18 \\
				FSCA (ours) & 67.40 & 65.91 & 72.10 &	68.86 &	74.28 &	16.70	 \\
				\bottomrule
			\end{tabular}
		\label{tab:LAION partial setting}
\end{table*}}

\textbf{LAION-mi in partial setting.} To conduct this experiment, for simplicity we use a partial setting of 50\% user knowledge~\footnote{In Sec~\ref{sec:ablation} we ablate different knowledge proportion in partial setting to show its influence on performance.}. We broadly consider existing auditing approaches designed for text-to-image
diffusion models as our baselines. We first consider Pixel Error~\cite{dubinski2024towards} and 
Attack-IV~\cite{wu2022membership} (the best-performing one among all variants in its paper) as they are both
black-box approaches. As shown in Tab~\ref{tab:LAION partial setting}, our FSCA yields superior results, outperforming them in all metrics by a large margin. For example, FSCA obtains 67.40\% Acc, outperforming the best black-box approach (Attack-IV) by around 7.7\%. To further demonstrate the effectiveness of our FSCA, we compare it with less restrictive gray-box approaches including Loss-based auditing~\cite{matsumoto2023membership}, SecMIstats (SecMI)~\cite{duan2023diffusion}, PIA~\cite{kongefficient},
PFAMI~\cite{fu2023probabilistic}, CLiD~\cite{zhai2024membership} and its Monte Carlo estimation version (M. C.). We find that our FSCA also is superior over all gray-box approaches. Considering the widely-used LAION-5b dataset in industry, the results of FSCA on LAION-mi dataset prove valuable for real-world auditing.


\setlength{\tabcolsep}{0.1cm}{\begin{table*}[htbp]
\centering
\vskip -0.1in
        \caption{Comparison of FSCA with previous approaches on \textbf{LAION-mi in shadow setting}. }
			\begin{tabular}{l c c c c c c}
				\toprule
				Approach & Acc$\uparrow$ & Pre$\uparrow$ & Rec$\uparrow$ &  F1$\uparrow$ & AUC$\uparrow$ & TPR@1\%FPR$\uparrow$ \\ 
				\midrule
                    \midrule
                    \multicolumn{4}{l}{\emph{Gray-box baseline}:}\\
                    Loss~\cite{matsumoto2023membership} & 51.77 & 51.95 & 47.30 & 49.51 & 53.36 & 1.15 \\
                    PIA~\cite{kongefficient} & 51.75  & 51.75 & 51.60 & 51.67 & 52.59 & 1.00  \\
                    SecMI~\cite{duan2023diffusion} & 51.60  & 51.16 & 70.45 & 59.27 &  52.98 & 0.70 \\
                    PFAMI~\cite{fu2023probabilistic} & 51.30 & 51.97 & 34.20 & 41.25 & 51.46 & 1.25  \\
                    M. C.~\cite{zhai2024membership} &  52.40 & 51.81 & 68.80 & 59.10 &   53.33 & 1.35  \\
    			  CLiD~\cite{zhai2024membership} &  57.08 & 56.97 & 57.91 & 57.43  & 60.58 & 3.05 \\
				\midrule 
    \multicolumn{4}{l}{\emph{Black-box baseline}:} \\
    Pixel Error~\cite{dubinski2024towards} & 50.45 & 50.91 & 25.15 & 33.67 & 49.49 & 0.85 \\
    Attack-IV~\cite{wu2022membership}  & 48.51 & 48.55 & 53.48 & 50.89 & 47.55 & 0.40  \\
				FSCA (ours) & 62.55 & 74.66 & 
  38.00 &	50.36 &	65.17 &	5.20 \\
				\bottomrule
			\end{tabular}
		\label{tab:LAION shadow setting}
\end{table*}}

\textbf{LAION-mi in Shadow Setting.} Besides partial setting, we conduct experiments on our shadow setting where the user has no knowledge of the internal training data used for the text-to-image diffusion system. Therefore, the user has to resort to public datasets. Here we use COCO dataset as the public dataset for simplicity. We train our shadow auditing model using FSCA framework on COCO dataset and then leverage this auditing model to evaluate on LAION-mi dataset. For baseline approaches, we follow the same procedure for fair comparison. We report the results in Tab~\ref{tab:LAION shadow setting}. It can be observed that the black-box approaches, Pixel Error and Attack-IV, both fail in auditing, whereas only our FSCA remains effective. For instance, our FSCA produces 62.55\% for accuracy while the accuracy of Attack-IV becomes 48.51\%.  Moreover, we find that though the performance of our FSCA decreases compared to partial setting, our approach still outperforms all gray-box approaches by a large margin across most metrics. Our FSCA outperforms the best-performing gray-box approach CLiD by around 5.5\% in accuracy. These results significantly demonstrates the effectiveness of our approach and show its potential in real-world auditing application.

\textbf{A Weaker Assumption in Shadow Setting.} Although in Tab~\ref{tab:LAION shadow setting} we assume that the user can access the entire image-text pairs used for auditing, we also report the performance of FSCA under a weaker assumption: the user only has images in hand without the corresponding text as we consider that this situation happens in reality, \eg, users sometimes upload their pictures on social media without including any descriptions. In this case, to employ our FSCA framework, we utilize the caption model BLIP-2~\footnote{We can also use other caption models, \eg, LLaVA~\cite{liu2023visual}. Since it is not the main focus of our article, we use BLIP-2 due to its superior performance on public dataset COCO.}~\cite{li2023blip} to supplement the missing text because it is well-trained on the COCO dataset, which may help us train our shadow auditing model. As presented in Tab~\ref{tab:LAION shadow setting weeker assumption}, even lacking text, our FSCA still outperforms all baselines across most metrics, demonstrating the potential of our approach.

\setlength{\tabcolsep}{0.1cm}{\begin{table*}[t]
\centering
\vskip -0.1in
        \caption{Comparison of FSCA with previous approaches on \textbf{LAION-mi in shadow setting with \textit{Pseudo Text}}. }
			\begin{tabular}{l c c c c c c}
				\toprule
				Approach (\textbf{\textit{Pseudo Text}}) & Acc$\uparrow$ & Pre$\uparrow$ & Rec$\uparrow$ &  F1$\uparrow$ & AUC$\uparrow$ & TPR@1\%FPR$\uparrow$ \\ 
				\midrule
                    \midrule
                    \multicolumn{4}{l}{\emph{Gray-box baseline}:}\\
                    Loss~\cite{matsumoto2023membership} & 52.50 &	52.76 &	47.30&	49.88 &	54.17 &	0.95 \\
                    PIA~\cite{kongefficient} &  51.63 &	51.62	& 51.75 &	51.69 &	52.49 &	0.75 \\
                    SecMI~\cite{duan2023diffusion} & 50.00 &	50.00 &	100.00	& 66.67 &	52.57 &	0.65 \\
                    PFAMI~\cite{fu2023probabilistic} &  50.75 &	51.09 &	34.08 &	40.89 &	50.88 &	1.15 \\
                    M. C.~\cite{zhai2024membership} &  52.60	& 51.94 &	69.52 &	59.46 &	53.01 &	1.25   \\
    			  CLiD~\cite{zhai2024membership} & 52.75 &	52.64 &	54.85 &	53.72 &	53.60 &	1.00 \\
				\midrule 
    \multicolumn{4}{l}{\emph{Black-box baseline}:} \\
    Pixel Error~\cite{dubinski2024towards} & 51.23 &	51.95 &	31.98 &	39.59 &	51.62 &	1.55 \\
    Attack-IV~\cite{wu2022membership}  & 44.15	& 44.66	 & 49.90	 & 47.14 &	43.92 &	0.40 \\
    FSCA & 58.73	& 73.51	& 27.04 &	39.54 &	61.72  &	8.15 \\
				\bottomrule
			\end{tabular}
		\label{tab:LAION shadow setting weeker assumption}
\end{table*}}

\setlength{\tabcolsep}{0.1cm}{\begin{table*}[htbp]
\centering
\vskip -0.1in
        \caption{Comparison of FSCA with previous approaches on \textbf{COCO in partial setting}.}
			\begin{tabular}{l c c c c c c}
				\toprule
				Approach & Acc$\uparrow$ & Pre$\uparrow$ & Rec$\uparrow$ &  F1$\uparrow$ & AUC$\uparrow$ & TPR@1\%FPR$\uparrow$ \\ 
				\midrule
                    \midrule
                    \multicolumn{4}{l}{\emph{Gray-box baseline}:}\\
				Loss~\cite{matsumoto2023membership} &  52.10 & 52.56 & 43.04 & 47.32 & 52.19 & 0.80  \\
                    PIA~\cite{kongefficient} &  54.08 & 54.59 & 48.46 & 51.35 & 55.52 & 1.76 \\
                    SecMI~\cite{duan2023diffusion} &  60.94 & 57.95 & 79.71 & 67.11 & 65.40  & 3.92  \\
                    PFAMI~\cite{fu2023probabilistic} & 57.36 & 55.92 & 69.54 & 61.99 & 60.39 & 2.72	 \\
                    M. C.~\cite{zhai2024membership} &  58.30 & 56.98 & 67.76 & 61.90 & 61.49 & 3.36 \\
    			  CLiD~\cite{zhai2024membership} & 88.42 & 89.15 & 87.50 & 88.31 & 95.47 & 58.35 \\
				\midrule 
    \multicolumn{4}{l}{\emph{Black-box baseline}:} \\
    Pixel Error~\cite{dubinski2024towards}  & 50.92 & 53.11 & 15.80 & 24.35 & 47.87 & 1.00 \\
    Attack-IV~\cite{wu2022membership} & 51.64 & 51.61 & 52.77 & 52.18 & 51.94 & 0.90 \\
				FSCA (ours) & 64.35 &	64.57 &	63.60 &	64.08 &	69.43 &	3.00  \\
				\bottomrule
			\end{tabular}
		\label{tab:coco partial setting}
\end{table*}}

\textbf{COCO in Partial Setting.} In addition of LAION-mi dataset, we further conduct experiments on COCO to verify the effectiveness and generalization of our FSCA framework. Similar to Tab~\ref{tab:LAION partial setting}, we also conduct a comprehensive comparison of our FACE with both black-box and gray-box baselines. The results are reported in Tab~\ref{tab:coco partial setting}. Although gray-box approaches typically have access to more internal information, FSCA still surpasses approaches like Loss, PIA, SecMI, PFAMI, and M.C. across most metrics. For example, FSCA produces 64.35\% for accuracy, outperforming SecMI by 3.4\%. These results demonstrate the great potential of our FSCA in auditing. At the same time, we observe that FSCA underperforms compared to CLiD on COCO, particularly in TPR@1\%FPR. Our findings suggest that CLiD's strong performance relies heavily on COCO's high-quality text-image alignment, as disrupting this alignment (via text shuffling~\cite{zhai2024membership}) causes CLiD's TPR@1\%FPR to plummet to 0.15—a ~99.8\% drop. This indicates that the membership feature constructed by CLiD is closely related to text-image alignment. This could also be a potential reason why CLiD performs poor on LAION-mi in Tab~\ref{tab:LAION partial setting} as LAION-mi relaxes the text-image alignment relatively. On the other hand, the competitive performance of FSCA on both COCO and LAION-mi demonstrates the robustness of our method under different text-image alignment conditions. Finally, while the performance of CLiD on COCO is remarkable, achieving this result necessitate the gray-box setting to provide CLiD with internal knowledge. Considering that reality is often more complex than CLiD assumes, CLiD may be not particularly easy to use.

\subsection{Effectiveness of Different Components in FSCA}\label{sec:ablation}

To reveal the effectiveness of the designs in FSCA, we conduct a series of ablation studies using variant models derived  our FSCA framework. Specifically, we verify the superiority of combining both semantic similarity and alignment compared with using them separately. We evaluate the performance of our FSCA under different proportions of knowledge assumptions in partial setting, \eg, 10\%, 20\%, \etc. We also compare the results of using queries of different number in our FSCA framework. We further report the performance when using different inference steps, which as shown in Fig~\ref{fig:demo-case} is an important parameter for diffusion generation. Finally we compare the performance of our two-branch auditing model with that of a potential one-branch model. We conduct all experiments in partial setting using the LAION-mi dataset, as it mitigates distributional mismatch, facilitating fairer comparisons with more meaningful results.

\setlength{\tabcolsep}{0.2cm}{\begin{table*}[htbp]
\centering
\vskip -0.1in
        \caption{Comparison of FSCA with previous approaches on LAION-mi in partial setting. \CheckmarkBold $\;$ indicates that the manner is employed.}
			\begin{tabular}{l c c c c c c c}
				\toprule
				Similarity & Alignment & Acc$\uparrow$ & Pre$\uparrow$ & Rec$\uparrow$ &  F1$\uparrow$ & AUC$\uparrow$ & TPR@1\%FPR$\uparrow$ \\ 
				\midrule
                    \midrule
                     \CheckmarkBold &  & 64.95	&68.12 &	56.20 &	61.59 &	69.32 &	16.4  \\
& \CheckmarkBold & 61.60 &	61.39 &	62.50 &	61.94	& 65.62 & 1.4  \\
				\CheckmarkBold & \CheckmarkBold & 67.40 & 65.91 & 72.10 &	68.86 &	74.28 &	16.7 \\
				\bottomrule
			\end{tabular}
		\label{tab:similarity and alignment}
\end{table*}}

\textbf{Semantic Similarity and Alignment.} Our FSCA involves two manners to capture the connection in text-image pairs: semantic similarity and semantic alignment. It is worth exploring how each manner performs. As shown in Tab~\ref{tab:similarity and alignment}, both semantic similarity and alignment yield satisfying results, even outperforming most baseline approaches. These results demonstrate that our adoption of the two manners are effective. Moreover, we find that semantic similarity is more important than semantic alignment according to the auditing performance. This result implies that semantic similarity, which measures the overall similarity in semantic between images and generated images, involves more crucial characteristic to capture the discrepancy between member and non-member. We speculate that this is because images contain more detailed or redundant information than text, resulting in finer-grained differences in content matching between real and generated images. Finally, when combining both of them, the performance is further improved, \eg, from 64.95\% Acc to 67.40\% Acc. This result indicates that the two manners are compatible.

\begin{figure}[htbp]
\centering\centerline{\includegraphics[width=1.0\linewidth]{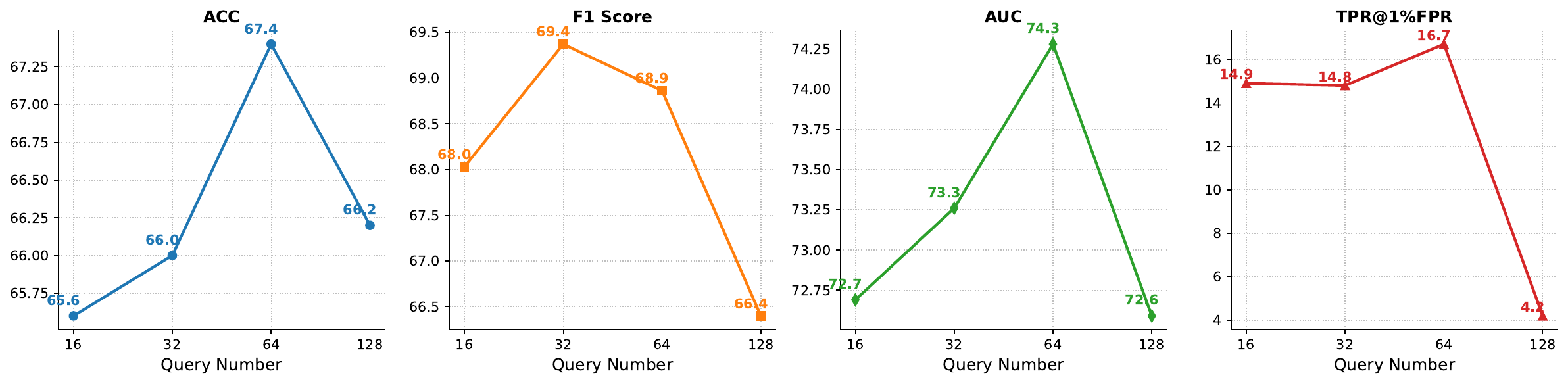}}
\caption{The impact of different query numbers on auditing performance in our FSCA.}
\label{fig:query_number}
\end{figure}

\textbf{Query Number.} In our FSCA framework, query number is an important hyperparameters and decides the dimension of input membership feature. In Fig~\ref{fig:query_number}, we consider four query numbers including 16, 32, 64, and 128. The experimental results show the impact of different query numbers on the FSCA framework's performance across ACC, F1 Score, AUC, and TPR@1\%FPR. One can see that,   as the query number increases, all metrics exhibit a similar trend: they gradually rise, reaching a peak at 64 queries, and then decline. We think that at the beginning, a higher number of queries provides the auditing model with more useful characteristics. However, beyond an optimal threshold, additional queries may introduce noise that misleads the auditing model, ultimately causing performance gains to plateau or even degrade. Moreover, considering that most text-to-image diffusion systems on the market are paid, a higher number of queries may impose a greater financial burden on users. Therefore, we use 64 as our default query number for account of both performance and financial burden.

\begin{figure}[t]
\centering\centerline{\includegraphics[width=1.0\linewidth]{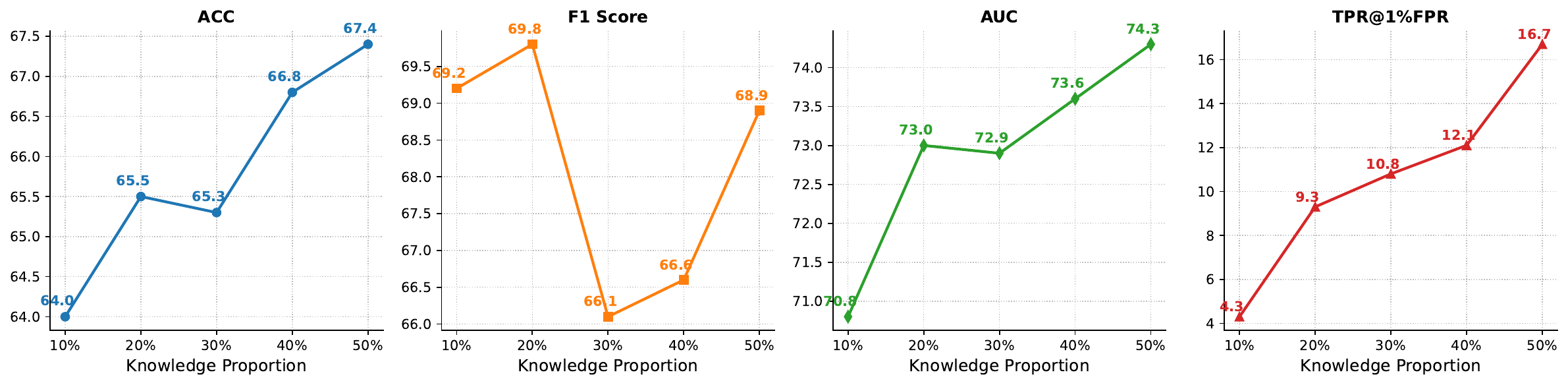}}
\caption{The impact of different knowledge proportions of training dataset on auditing performance in our FSCA.}
\label{fig:knowledge proportions}
\end{figure}

\textbf{Knowledge Proportion.} In our partial setting, we use 50\% proportion of training dataset by default as our user prior knowledge. However, this proportion varies widely in real-world scenarios. In this experiment, we consider five different proportion settings ranging from 10\% to 50\%, and evaluate the impact of these knowledge propositions on the performance of our FSCA framework using ACC, F1 Score, AUC, and TPR@1\%FPR. The results are illustrated in Fig~\ref{fig:knowledge proportions}. One can see that as the knowledge proposition increases, there is a consistent improvement across most metrics, \eg, ACC (64.0\% to 67.4\%), AUC (70.8\% to 74.3\%), and TPR@1\%FPR (4.3 to 16.7), indicating that the knowledge propositions is important to enhance the model's predictive performance. This suggests that users should gather as much internal training data as possible to enhance auditing reliability. Moreover, compared to other metrics, the TPR@1\%FPR metric exhibits remarkable variability (ranging from 4 to 16), suggesting that the impact of knowledge proportions on this specific aspect is significant. Overall, these results highlight the positive influence of knowledge propositions on the framework's performance.

\begin{figure}[htbp]
\centering\centerline{\includegraphics[width=1.0\linewidth]{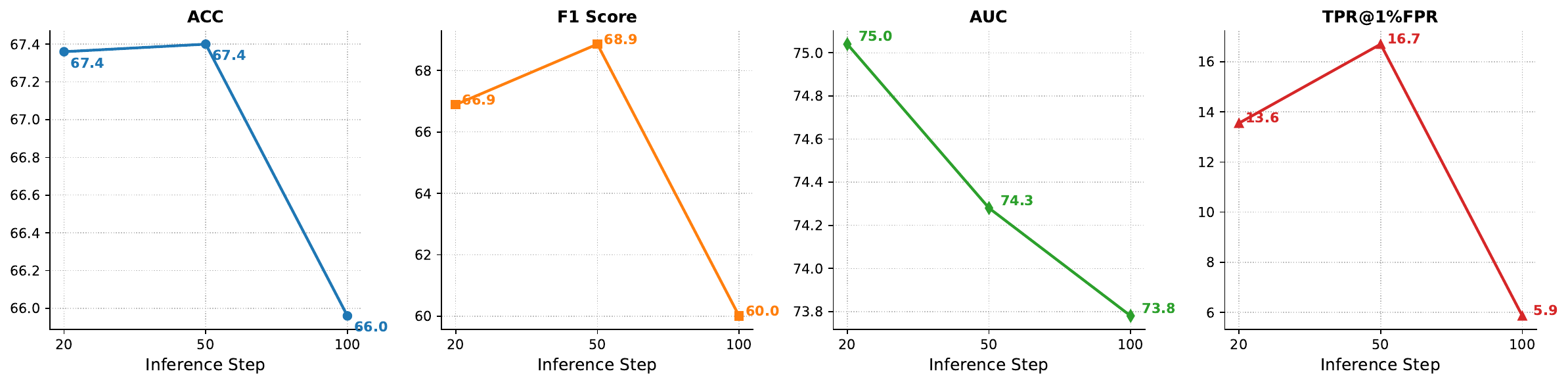}}
\caption{The impact of different inference steps during generation on auditing performance in our FSCA.}
\label{fig:inference steps}
\end{figure}

\textbf{Inference Step.} In this experiment, we investigate the effect of varying inference steps (20, 50, and 100), which as shown in Fig~\ref{fig:demo-case} is an important parameter for diffusion generation, on auditing performance across ACC, F1 Score, AUC, and TPR@1\%FPR. As presented in Fig~\ref{fig:inference steps}, the initial improvement in metrics such as F1 Score (from 66.9\% to 68.9\%) and TPR@1\%FPR (from 13.6 to 16.7) suggests that more inference steps allow the model to refine its predictions, leading to better precision-recall balance and higher true positive rates at low false positive rates. However, the decline in performance when the number of inference steps is increased further (e.g., ACC dropping from 67.4\% to 66.0\% and TPR@1\%FPR decreasing from 16.7 to 5.9) indicates that there is a point of diminishing returns. Beyond this point, the additional computational effort does not translate into better performance. This may be because the model over-refines the outputs during the diffusion process, thereby weakening the discrepancy between members and non-members. This experiment highlights the importance of inference step and suggests that users could tune this parameter carefully to achieve the best auditing performance.

\setlength{\tabcolsep}{0.2cm}{\begin{table*}[t]
\centering
        \caption{Comparison of FSCA with previous approaches on LAION-mi in partial setting. }
			\begin{tabular}{l c c c c c c c}
				\toprule
				Approach  & Acc$\uparrow$ & Pre$\uparrow$ & Rec$\uparrow$ &  F1$\uparrow$ & AUC$\uparrow$ & TPR@1\%FPR$\uparrow$ \\ 
				\midrule
                    \midrule
                    
 One Branch & 65.55 & 67.45 & 60.10 &	63.56 &	73.71 &	9.2 \\
				 Two Branches (FSCA) & 67.40 & 65.91 & 72.10 &	68.86 &	74.28 &	16.7 \\
				\bottomrule
			\end{tabular}
		\label{tab:auditing structure}
        \vskip -0.1in
\end{table*}}

\textbf{Auditing Model Structure.} In our FSCA framework, we use two-branches auditing model. In fact, there is also an one-branch alternative. As shown in Fig~\ref{fig:overview} (d), our two-branch auditing model first  processes the two types of membership features independently and then concatenate the outputs from align stream and similarity stream. In contrast, the alternative one-branch auditing model first concatenate the two types of membership features and then processes them. The difference lies in the processing sequence. As shown in Tab~\ref{tab:auditing structure}, we find that the two-branches structure performs better than the one-branch one across most metrics. For example, two-branches structure produces 67.40\% for accuracy, outperforming one-branch one by around 2\%. Moreover, we find two-branches structure significantly outperforms its counterpart in TPR@1\%FPR by around 7.5. 
We speculate that directly combining the two types of membership features may introduce some incompatibility, ultimately compromising auditing performance. Hence, in our FSCA framework, we leverage the two-branches structure to yield better auditing performance. It is also interesting to explore how to design the one-branch alternative to catch even surpass two-branches one, which we leave as future work.

\begin{figure}[htbp]
\centering\centerline{\includegraphics[width=1.0\linewidth]{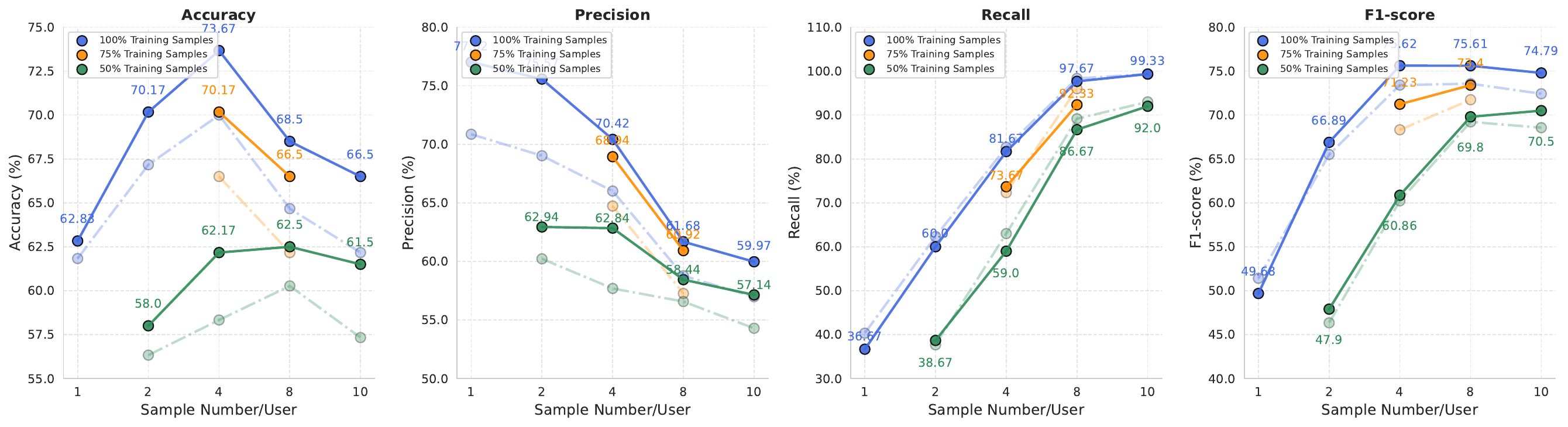}}
\caption{The performance of a user-level auditing in shadow setting when different proportions of training samples are included in this user's auditing samples.}
\label{fig:user level stage1}
\end{figure}

\subsection{A User-level Auditing in Real-world Scenarios}

Let's consider a real-world case where a user who have many data samples, \ie, text-image pairs in her social media, wants to know whether her samples are used for training a text-to-image diffusion system. In this context, instead of auditing individual sample, we focus on user-level auditing where the user’s (some) data samples are used—possibly in combination with the data samples from thousands of other users—to train the model. Particularly, if we can determine that any single sample is used for training, we can conclude that the user's data has been utilized for training~\cite{song2019auditing}. On the other hand, if none of the user's samples are used for training, the user is fortunate. To show the performance of our FSCA framework in this situation, we consider 100 \textit{victim users} who have some data samples with a certain proportion involved in training and 100 \textit{fortunate users} whose data samples are not. Our goal is to test how many users (both victim and fortunate) are identified under our FSCA framework. We conduct all experiments in shadow setting and repeat three times to make our results more practically valuable and illustrate the averaged auditing results in Fig~\ref{fig:user level stage1}. The x-axis represents the number of samples a user may have, while the y-axis reports the user-level accuracy, precision, recall and F1-score. The values (100\%, 75\%, and 50\%) means the proportion of a user's data samples are involved in training.


\begin{figure}[htbp]
\centering\centerline{\includegraphics[width=1.0\linewidth]{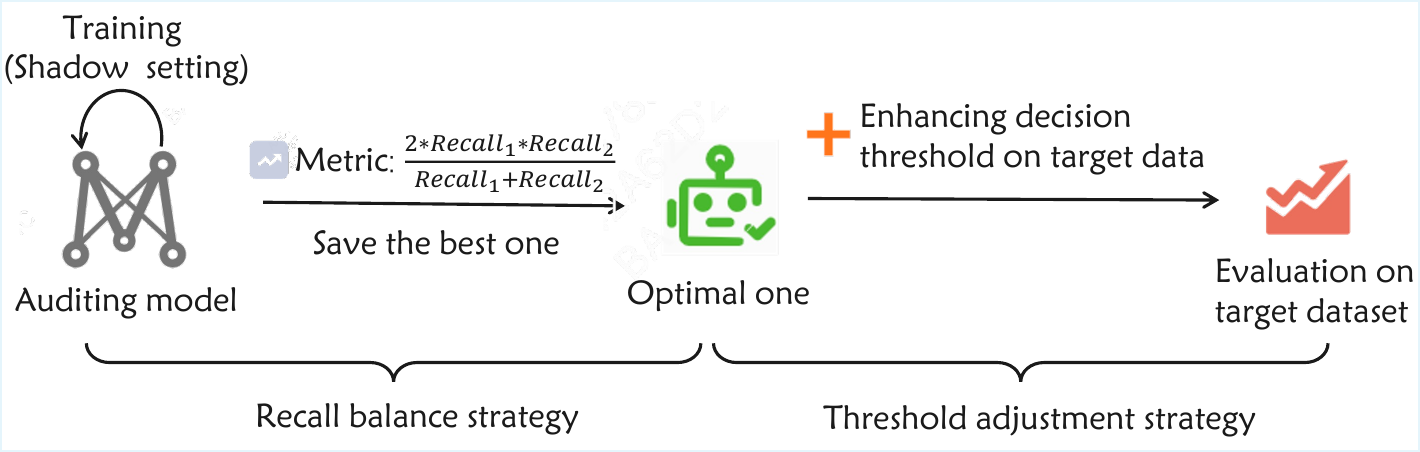}}
\caption{The illustrated process. $Recall_1$ represents recall of of member samples and $Recall_2$ represents recall of non-member samples.}
\label{fig:process}
\end{figure}

In Fig~\ref{fig:user level stage1}, our initial experiments, represented by the light-colored curves, reveal unsatisfactory performance. This result is due to the increasing difficulty in accurately predicting \textit{fortunate users} as the sample number increases. Specifically, if a fortunate user is correctly identified, all her samples should be classified as non-member. Therefore, if sample number is $n$ and the accuracy of an auditing model is $A$ ($A<1$), the final success probability should be $A^{n}$. As $n$ increases, the success probability declines, leading to severe classification errors. Consequently, we can see that the performance eventually deteriorates as the sample number grows. From the auditing model's perspective, this issue is a result of low recall of non-member samples, where non-member samples are incorrectly classified as member samples, leading to misidentification of fortunate users as victim users. Therefore we can improve recall for these non-member samples. Simultaneously, it's also crucial to consider the recall of member samples, as it directly impacts victim user predictions. To strike a balance between these two aspects, we heuristically propose a recall balance strategy~\footnote{We also consider other metrics, \eg, precision and F1, in our preliminary experiments. However, we find that the performance is inferior to that of our recall balance strategy.} as shown in Fig~\ref{fig:process} (left part), which adopts an F1-score-based formulation incorporating recall for both member and non-member samples and saves the auditing model that achieves the highest score for evaluation. When employing the strategy, compared with the light-colored curves, the performance, represented by the dark-colored curves, is significantly improved at the cost of slightly decreased recall, demonstrating the effectiveness of this strategy.

\begin{figure}[t]
\centering\centerline{\includegraphics[width=1.0\linewidth]{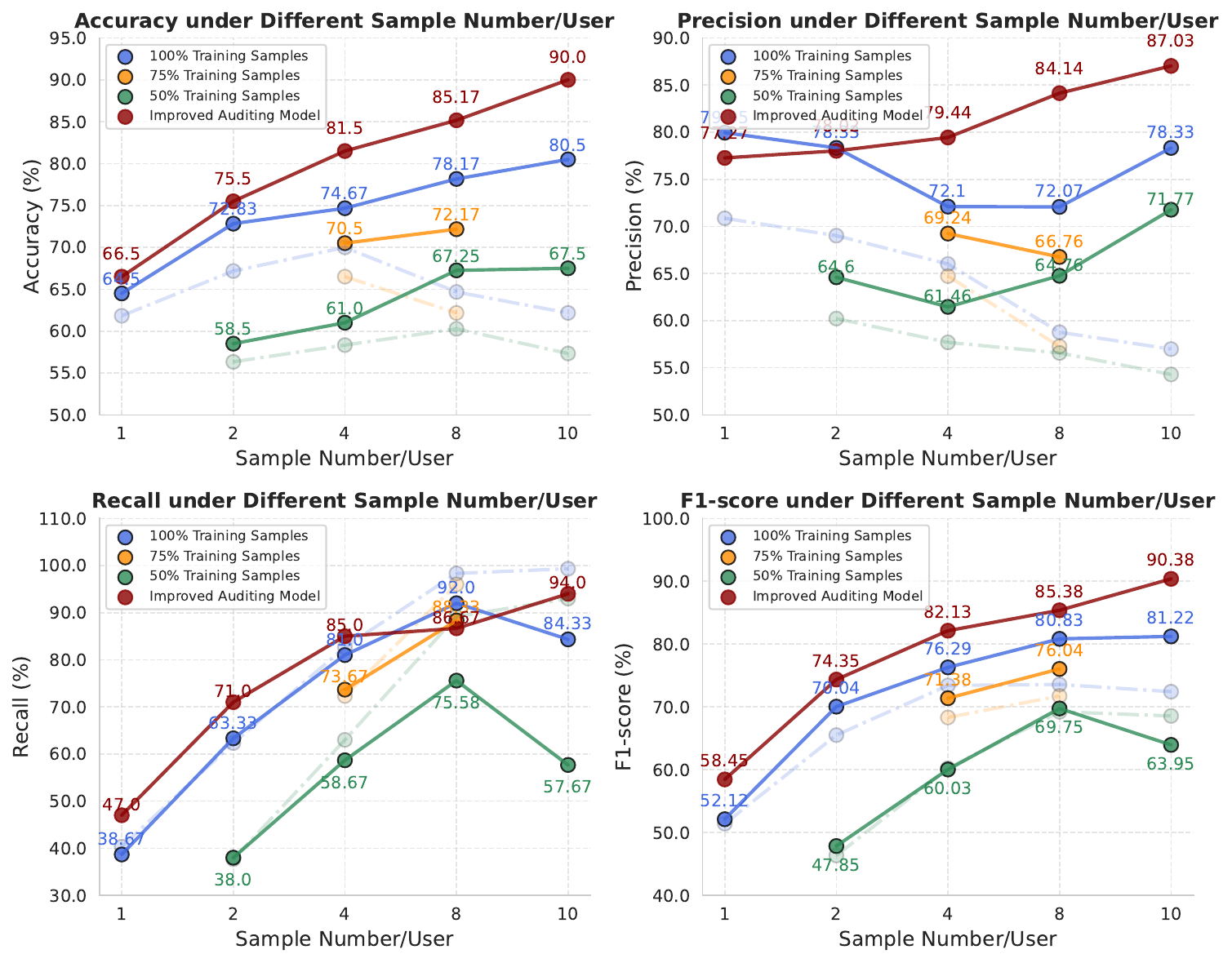}}
\caption{The user-level auditing performance comparison when employing threshold adjustment strategy (represented by dark-colored curves). The light-colored curves represent initial experimental results.}
\label{fig:user level}
\end{figure}

In addition, we propose another simple yet effective strategy called threshold adjustment strategy as shown in Fig~\ref{fig:process} (right part) to further improve auditing performance. Our insight is that the auditing model typically exhibits low confidence when incorrectly predicting a non-member target sample (\ie, from a fortunate user) as a member. Hence, setting a relatively large confidence threshold helps prevent misidentifying a non-member target sample from the fortunate user as a member sample. To fully illustrate the potential of this strategy, we conduct a detailed grid search (0.01) of threshold during the evaluation of the target dataset. For example, we set threshold to 0.53 when sample number is 4~\footnote{We give a detailed threshold setting for different sample number in Sec~\ref{sec:hyper}}. The results are reported in Fig~\ref{fig:user level}. One can see that this strategy consistently enhances the performance at the cost of slightly decreased recall. Moreover, we observe that when the sample number is fixed, performance improves as the proportion of training samples increases. For example, when the sample number is 4 and proportion increases from 50\% to 75\%, the accuracy improves from 61\% to 70.5\%. This suggests that users whose samples are incorporated into training at a higher proportion are easier to audit. Finally, we observe that using just 10 samples allows our framework to achieve up to 80\% accuracy, implying the promising potential of FSCA.

Finally, to further improve the auditing performance, we combine recall balance strategy with threshold adjustment strategy. We illustrate the whole process in Fig~\ref{fig:process} and present the results in Fig~\ref{fig:user level} (red curves) in the proportion setting of 100\% training samples. As shown, this combination leads to consistent improvements across all metrics. Notably, our approach achieves 90\% accuracy with 10 samples/user, highlighting the practical value of our framework in real-world auditing scenarios.

\section{Ethical and Societal Impact}

In this study, we discuss the ethical and societal implications of auditing user data to determine its usage in the training of text-to-image diffusion models. Our analysis intends to identify potential misuse and privacy concerns associated with the unauthorized use of personal data in these models. We propose a black-box and realisticly effective auditing framework called FSCA to enhance transparency and accountability in the data collection process, which will encourage AI service providers to proactively address the challenges related to data privacy and security, ensuring that user data is handled ethically and responsibly. We believe that our research further advances the development of data governance in AI application and the promotion of ethical practices and privacy protection, ultimately fostering a safer and more reliable digital ecosystem. However, despite our intention to foster ethical data usage and promote responsible AI research, there is a possibility that the auditing tools could be misused by certain malicious people to infer user data privacy. This highlights the need for collective vigilance and supervision by all stakeholders involved.

\section{Methods}

This section  provides a brief introduction of text-to-image diffusion models and membership auditing, followed by a comprehensive overview of the workflow
 of our framework. We will describe the specific function of each part and detail the hyperparameters and configurations during our experiments.

\subsection{Preliminaries}


\subsubsection{Text-to-image Diffusion Models}
Text-to-image diffusion models are grounded in the framework of denoising diffusion probabilistic models (DDPM)~\cite{ho2020denoising} and leverage conditional modeling to align textual inputs with image synthesis. This subsection provides an overview of the fundamental principles underlying these models, including the diffusion process, conditional generation, and training objectives.

\begin{figure}[t]
\centering\centerline{\includegraphics[width=1.0\linewidth]{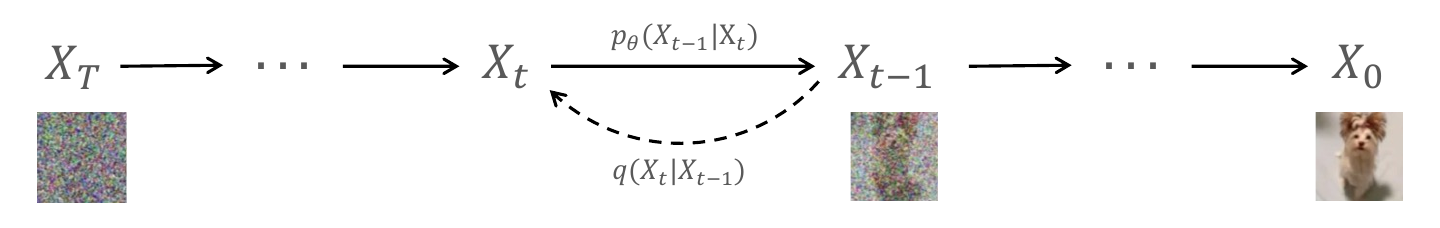}}
\caption{An illustration of diffusion and denoising process in the framework of denoising diffusion probabilistic models (DDPM)~\cite{ho2020denoising}.}
\label{fig:ddpmcase}
\end{figure}

\textbf{Diffusion Process.} As illustrated in Fig~\ref{fig:ddpmcase}, diffusion models operate by transforming a data distribution $p(\mathbf{x}_0)$ into a Gaussian distribution through a forward diffusion process (dashed arrow in Fig~\ref{fig:ddpmcase}) and learning the reverse denoising process (solid arrow in Fig~\ref{fig:ddpmcase}). The forward process $q(\mathbf{x}_t | \mathbf{x}_{t-1})$ progressively adds Gaussian noise at each step $t$:
\begin{equation}
q(\mathbf{x}_t | \mathbf{x}_{t-1}) = \mathcal{N}(\mathbf{x}_t; \sqrt{1 - \beta_t} \mathbf{x}_{t-1}, \beta_t \mathbf{I}),
\end{equation}
where $\beta_t \in (0, 1)$ is a variance schedule controlling the noise level. After $T$ steps, the data distribution $p(\mathbf{x}_0)$ is transformed into an isotropic Gaussian $p(\mathbf{x}_T) \sim \mathcal{N}(0, \mathbf{I})$.

The reverse process, parameterized by a neural network, learns to denoise $\mathbf{x}_t$ iteratively:
\begin{equation}
p_\theta(\mathbf{x}_{t-1} | \mathbf{x}_t) = \mathcal{N}(\mathbf{x}_{t-1}; \mu_\theta(\mathbf{x}_t, t), \Sigma_\theta(\mathbf{x}_t, t)),
\end{equation}
where $\mu_\theta$ and $\Sigma_\theta$ are predicted by the model.

\textbf{Conditional Diffusion for Text-to-Image Generation.} In text-to-image tasks, the reverse process incorporates textual conditions $\mathbf{c}$ to guide image synthesis. The conditional reverse process is defined as:
\begin{equation}
p_\theta(\mathbf{x}_{t-1} | \mathbf{x}_t, \mathbf{c}) = \mathcal{N}(\mathbf{x}_{t-1}; \mu_\theta(\mathbf{x}_t, t, \mathbf{c}), \Sigma_\theta(\mathbf{x}_t, t, \mathbf{c})).
\end{equation}

To represent textual inputs, pretrained language models (e.g., CLIP text encoder~\cite{radford2021learning} or T5 text encoder~\cite{raffel2020exploring}) are used to embed the text into a latent space:
\begin{equation}
\mathbf{c} = \text{TextEncoder}(\text{text input}),
\end{equation}
where $\mathbf{c}$ is subsequently integrated with the denoising network via mechanisms such as cross-attention~\cite{vaswani2017attention}.

\textbf{Training Objective.} The training objective minimizes the reconstruction error between the predicted noise $\epsilon_\theta$ and the actual noise $\epsilon$ added in the forward process. The simplified loss function is:
\begin{equation}
\mathcal{L} = \mathbb{E}_{\mathbf{x} \sim p(\mathbf{x}_0), \epsilon \sim \mathcal{N}(0, \mathbf{I}), t} \left[ \|\epsilon - \epsilon_\theta(\mathbf{x}_t, t, \mathbf{c}) \|^2 \right].
\end{equation}

This objective ensures that the model learns to denoise effectively, progressively reconstructing the original data from noisy intermediate states.

\subsubsection{Membership Auditing} \label{sec:membership inference}

The problem of membership auditing is closely related to the problem of membership inference (MI)~\cite{hu2022membership, shokri2017membership}. This subsection provides a problem definition of membership auditing and then clarifies the relationship between membership auditing and MI.

\textbf{Problem Definition.} The goal of a membership auditing is to determine whether a given data point $\mathbf{x}$ is part of a model's training dataset $\mathcal{D}_\text{train}$. Formally, given a target model $f_\theta$ trained on $\mathcal{D}_\text{train}$, the user constructs a classifier $h$ that outputs:
\begin{equation}
h(\mathbf{x}, f_\theta) = 
\begin{cases} 
1, & \text{if } \mathbf{x} \in \mathcal{D}_\text{train}, \\ 
0, & \text{otherwise.}
\end{cases}
\end{equation}

For example, the classifier $h$ can leverages model outputs like confidence scores~\cite{salem2018ml} to audit membership status.

\textbf{Relevance to Membership Inference.} Membership auditing and  membership  inference (MI) are inherently two sides of the same coin, distinguished primarily by the intent of their application. By probing whether a specific data point is included in a model's training dataset, MI reveals potential risks of overfitting and unintended data leakage. However, from the perspective of users, MI could serve as a crucial tool for privacy auditing, helping users to determine whether their data has been utilized for training models without their authorization. This duality underscores the importance of MI as both a potential vulnerability in machine learning and a valuable tool for the public to protect sensitive information and intellectual property. In the following, we use the terms ``auditing" or ``membership auditing" for simplicity, as we are utilizing this technology for positive purposes.

\subsection{Overview}

Our Feature Semantic Consistency-based Auditing (FSCA) framework contains four stages including fake image generation (Fig~\ref{fig:overview} (a)), feature extraction (Fig~\ref{fig:overview} (b)), membership feature generation (Fig~\ref{fig:overview} (c)), and auditing model training (Fig~\ref{fig:overview} (d)). 

\textbf{Stage 1: Fake Image Generation.} Given a image-text pair, we first feed the text into the text-to-image diffusion system to generate $N$ fake images. These images are regarded as queries for next stage. It is worth noting that in this stage, we do not have access to the internal information or structure of this diffusion system. This guarantees that our FSCA remains a fully black-box approach, enabling its practical application.

\textbf{Stage 2: Feature Extraction.} After obtaining generated images, we extract (generated) images and text features via
mapping them to representation space through an image 
encoder and a text encoder, respectively. Note that we use multimodally trained image encoder and text encoder, \eg, CLIP~\cite{radford2021learning}, and leverage the same image encoder for the real image and generated images. By doing so, we can ensure the text feature, image feature, and generated image feature are all embedded in the same representation space, making following comparison reasonable and meaningful.

\textbf{Stage 3: Membership Feature Generation.} Essentially, training a text-to-image diffusion model is to bridge the connection in text-image pair and model their joint distribution. In our FSCA, we consider two manners to capture the connection including semantic alignment and semantic similarity. To capture the semantic alignment, we compute the alignment score between text feature and generated image features using cosine similarity. We denote these scores as $[\alpha_{1}, \alpha_{2}, ..., \alpha_{N}]$. Similarly, we compute an alignment base score $[\alpha^{b}]$ using text feature and image feature. To capture semantic similarity, we compute the cosine similarity between image feature and generated image features. These similarity scores are denoted as $[s_{1}, s_{2}, ..., s_{N}]$. Together, we refer to the resulting scores as membership features, which serve as input for subsequent stage.

\textbf{Stage 4: Auditing Model Training.} Before training an auditing model, we need process the obtained input from last stage. For alignment scores, they quantify the degree of alignment between text and generated images. Ideally, if a text-image pair is included in training, these scores should closely approximate the alignment base score. Therefore, we subtract $a^{b}$ from each alignment score, yielding a metric called the alignment difference score, which serves as the input to our auditing model.  For similarity scores, they directly indicate the degree of correspondence between an image and its generated counterparts. Intuitively, if a text-image pair is part of the training dataset, these scores should be relatively high. To the end, we take alignment difference score and similarity score as model input. Inspired by previous work~\cite{zhu2022safety, zhu2024safety}, our auditing model includes three parts: align stream, similarity stream, and fusion stream. The align stream processes alignment difference score. The similarity stream processes similarity score. Then, the fusion stream fuses the features extracted from the above two streams and outputs a probability to indicate whether the
text-image pair is used in training or not.

\subsection{Hyperparameters and Configurations}~\label{sec:hyper}

When training our auditing model, we follow previous approaches~\cite{zhu2022safety, zhu2024safety} and initialize all the network weights with normal distribution, and all biases with 0 by default. We use Tanh as the activation function between two fully connected layers. We set the batch size to 100, use the Adam optimizer with the
learning rate of 0.001 and weight decay set to 0.0005. Following
previous studies~\cite{nasr2018machine, zhu2024unified, liu2022ml, zhu2025unified}, we train our auditing models for 100 epochs. Moreover, during
the training process, we ensure that every training batch contains the same number of member and non-member data samples. By doing so, we prevent the auditing model from being biased toward either side.

In partial setting, for experiments on COCO dataset, we follow previous work~\cite{zhai2024membership} and select 2500/2500 members/non-members from MS-COCO. We use text-to-image diffusion model, Stable Diffusion v1-4~\footnote{\url{https://huggingface.co/CompVis/stable-diffusion-v1-4}} (SD v1.4) and follow previous work~\cite{zhai2024membership} to fine-tune it with 50,000 steps using 2500 members with a 1e-5 learning rate and 4 batch size. Additionally, we employ the default data augmentation (Random-Crop and Random-Flip) in the training~\footnote{\url{https://github.com/huggingface/diffusers/blob/main/examples/text_to_image/train_text_to_image.py}}. The fine-tuned model is considered as our target model. For LAION-mi dataset, we follow previous work~\cite{zhai2024membership} and conduct auditing experiments on Stable Diffusion v1-5~\footnote{\url{https://huggingface.co/CompVis/stable-diffusion-v1-5}} directly as SD v1.5 is trained on LAION-mi member set. Similarly, we select 2500/2500 members/non-members from LAION-mi dataset. When it comes to auditing, for both COCO and LAION-mi, we use pre-defined knowledge proportion (\eg, 50\%) of the dataset to train our auditing model and adopt the remaining data for evaluation.

In shadow setting, for experiments on LAION-mi, we first use 2500/2500 members/non-member of COCO dataset, which could be considered as public dataset, to train our auditing model on fine-tuned SD v1.4. Then we evaluate the performance of this auditing model on LAION-mi that could be considered as a private dataset. To compare with the results in partial setting, we still use the remaining data of LAION-mi when knowledge proportion is 50\% for evaluation.

For baseline approaches, we use the parameters recommended in their papers. When auditing, we follow the training and evaluation procedure of both partial setting and shadow setting for fair comparisons.

For the user-level auditing experiments, we set threshold to 0.5 for light-colored curves by default. For dark-colored curves, we set threshold to 0.52 when sample number is 1 and 2. We set threshold to 0.53 and 0.56 when sample number is 4 and 8, respectively. We set threshold to 0.61 when sample number is 10.

\section{Data availability}
COCO dataset and LAION-mi are all derived from
previous studies. Here we include the official links. COCO: \url{https://huggingface.co/datasets/zsf/COCO_MIA_ori_split1}. LAION-mi: \url{https://drive.google.com/drive/folders/17lRvzW4uXDoCf1v_sIiaMnKGIARVunNU}. COCO in our experiments is derived from MS COCO: \url{https://cocodataset.org/}. LAION-mi is derived from LAION-5b: \url{https://laion.ai/blog/laion-5b/}.

\section{Code availability}

We built FSCA using Python and the Pytorch deep learning
framework. The code repository of FSCA, readme files and tutorials
are all available at \url{https://github.com/JiePKU/FSCA}.

\bibliography{sn-bibliography} 

\end{document}